\documentclass{article}
\usepackage{graphics,graphicx,caption,float,subcaption,booktabs,xcolor,multirow,array,color,ifthen,tabu,colortbl,dblfloatfix,url}
\usepackage{algorithmic,amssymb,xspace,nicefrac,microtype}
\usepackage{amsmath,amsfonts,bm}
\usepackage[sort]{natbib}
\usepackage[pagebackref=true,breaklinks=true,colorlinks=true,bookmarks=false]{hyperref}

\usepackage[accepted]{icml2019}

\icmltitlerunning{Self-Supervised Exploration via Disagreement}
\begin{document}

\twocolumn[
\icmltitle{Self-Supervised Exploration via Disagreement}

\icmlsetsymbol{equal}{*}
\begin{icmlauthorlist}
\icmlauthor{Deepak Pathak}{equal,ucb}
\icmlauthor{Dhiraj Gandhi}{equal,cmu}
\icmlauthor{Abhinav Gupta}{cmu,fair}
\end{icmlauthorlist}
\icmlaffiliation{ucb}{UC Berkelely}
\icmlaffiliation{cmu}{CMU}
\icmlaffiliation{fair}{Facebook AI Research}
\icmlcorrespondingauthor{Deepak Pathak}{pathak@cs.berkeley.edu}
\icmlkeywords{exploration, curiosity, sparse reward, robot, efficient, reinforcement learning, model learning, differentiable exploration}
\vskip 0.3in
]
\printAffiliationsAndNotice{\icmlEqualContribution}

\begin{abstract}
Efficient exploration is a long-standing problem in sensorimotor learning. Major advances have been demonstrated in noise-free, non-stochastic domains such as video games and simulation. However, most of these formulations either get stuck in environments with stochastic dynamics or are too inefficient to be scalable to real robotics setups. In this paper, we propose a formulation for exploration inspired by the work in active learning literature. Specifically, we train an ensemble of dynamics models and incentivize the agent to explore such that the disagreement of those ensembles is maximized. This allows the agent to learn skills by exploring in a self-supervised manner \textit{without any external reward}. Notably, we further leverage the disagreement objective to optimize the agent's policy in a \textit{differentiable manner}, without using reinforcement learning, which results in a sample-efficient exploration. We demonstrate the efficacy of this formulation across a variety of benchmark environments including stochastic-Atari, Mujoco and Unity. Finally, we implement our differentiable exploration on a real robot which learns to interact with objects completely from scratch. Project videos and code are at~\url{https://pathak22.github.io/exploration-by-disagreement/}.
\end{abstract}

\section{Introduction}
\label{sec:intro}
Exploration is a major bottleneck in both model-free and model-based approaches to sensorimotor learning. In model-based learning, exploration is a critical component in collecting diverse data for training the model in the first place.
On the other hand, exploration is indispensable in model-free reinforcement learning (RL) when rewards extrinsic to the agent are sparse.
The common approach to exploration has been to generate ``intrinsic'' rewards, i.e., rewards automatically computed based on the agents model of the environment.
Existing formulations of intrinsic rewards include maximizing ``visitation count''~\cite{bellemare2016unifying,poupart2006analytic,lopes2012exploration} of less-frequently visited states, ``curiosity''~\cite{pathakICMl17curiosity,oudeyer2009intrinsic,schmidhuber1991curious} where prediction error is used as reward signal and ``diversity rewards''~\cite{eysenbach2018diversity,lehman2011evolving,lehman2011abandoning} which incentivize diversity in the visited states. These rewards provide continuous feedback to the agent when extrinsic rewards are sparse, or even absent altogether.

Generating intrinsic rewards requires building some form of a predictive model of the world. However, there is a key challenge in learning predictive models beyond noise-free simulated environments: how should the stochastic nature of agent-environment interaction be handled? Stochasticity could be caused by several sources: (1) noisy environment observations (e.g, TV playing noise), (2) noise in the execution of agent's action (e.g., slipping) (3) stochasticity as an output of the agent's action (e.g., agent flipping coin). One straightforward solution to learn a predictive forward model that is itself stochastic!
Despite several methods to build stochastic models in low-dimensional state space~\cite{chua2018deep,houthooft2016vime}, scaling it to high dimensional inputs (e.g., images) still remains challenging.
An alternative is to build deterministic models but encode the input in a feature space that is invariant to stochasticity.
Recent work proposed building such models in inverse model feature space~\cite{pathakICMl17curiosity} which can handle stochastic observations but fail when the agent itself is the source of noise (e.g. TV with remote~\cite{burda2018large}).

\begin{figure*}[t]
\centering
\includegraphics[width=0.84\linewidth]{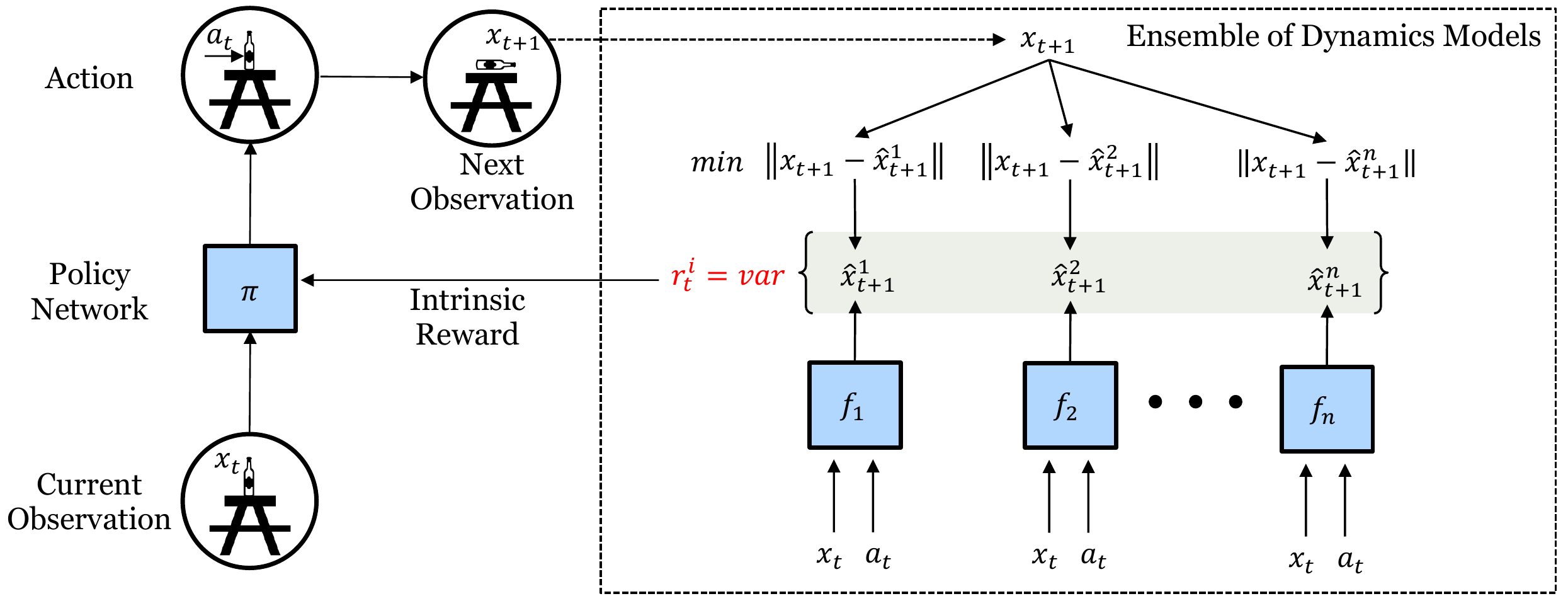}
\vspace{-0.1in}
\caption{\textbf{Self-Supervised Exploration via Disagreement}: At time step $t$, the agent in the state $x_{t}$ interacts with the environment by taking action $a_{t}$ sampled from the current policy $\pi$ and ends up in the state $x_{t+1}$. The ensemble of forward models $\{f_{1},f_{2},...,f_{n}\}$ takes this current state $x_t$ and the executed action $a_t$  as input to predict the next state estimates $\{\hat x^{1}_{t+1}, \hat x^{2}_{t+2},...,\hat x^{n}_{t+1}\}$. The variance over the ensemble of network output is used as intrinsic reward $r^{i}_{t}$ to train the policy $\pi$. In practice, we encode the state $x$ into an embedding space $\phi (x)$ for all the prediction purposes.}
\label{fig:method}
\vspace{-0.1in}
\end{figure*}

Beyond handling stochasticity, a bigger issue in the current intrinsic reward formulations is that of sample efficiency. The agent performs an action and then computes the reward based on its own prediction and environment behavior. For instance, in curiosity~\cite{pathakICMl17curiosity,oudeyer2009intrinsic}, the policy is rewarded if the prediction model and the observed environment disagree. From an exploration viewpoint, this seems like a good formulation, i.e, rewarding actions whose effects are poorly modeled. But this reward is a function of environment dynamics with respect to the performed action. Since the environment dynamics is unknown, it is treated as black-box and the policy's gradients have to be estimated using high-variance estimators like REINFORCE~\cite{williams1992simple} which are extremely sample-inefficient in practice.

We address both the challenges by proposing an alternative formulation for exploration taking inspiration from active learning. The goal of active learning is to selectively pick samples to label such that the classifier is maximally improved.
However, unlike current intrinsic motivation formulations where an agent is rewarded by comparing the prediction to the ground-truth, the importance of a sample is not computed by looking at the ground-truth label but rather by looking at the state of the classifier itself.
For instance, a popular approach is to label the most uncertain samples by looking at the confidence of the classifier.
However, since most of the high-capacity deep neural networks tend to overfit, confidence is not a good measure of uncertainty.
Hence, taking an analogy from the Query-by-Committee algorithm~\cite{Seung1992}, we propose a simple disagreement-based approach: we train an ensemble of forward dynamics models and incentivize the agent to explore the action space where there is maximum disagreement or variance among the predictions of models of this ensemble.
Taking actions to maximize the model-disagreement allows the agent to explore in a completely self-supervised manner without relying on any external rewards.
We show that this approach does not get stuck in stochastic-dynamics scenarios because all the models in the ensemble converge to mean, eventually reducing the variance of the ensemble.

Furthermore, we show that our new objective is a differentiable function allowing us to perform policy optimization via direct likelihood maximization -- much like supervised learning instead of reinforcement learning. This leads to a sample efficient exploration policy allowing us to deploy it in a real robotic object manipulation setup with 7-DOF Sawyer arm. We demonstrate the efficacy of our approach on a variety of standard environments including stochastic Atari games~\cite{stickyAtari}, MNIST, Mujoco, Unity~\cite{unity_ml} and a real robot.

\section{Exploration via Disagreement}
Consider an agent interacting with the environment $\mathcal{E}$. At time $t$, it receives the observation $x_t$ and then takes an action predicted by its policy, i.e., $a_t \sim \pi(x_t; \theta_P)$. Upon executing the action, it receives, in return, the next observation $x_{t+1}$ which is `generated' by the environment. Our goal is to build an agent that chooses its action in order to maximally explore the state space of the environment in an efficient manner.
There are two main components to our agent: an intrinsic forward prediction model that captures the agent's current knowledge of the states explored so far, and policy to output actions.
As our agent explores the environment, we learn the agent's forward prediction model to predict the consequences of its own actions.
The prediction uncertainty of this model is used to incentivize the policy to visit states with maximum uncertainty.

Both \textit{measuring} and \textit{maximizing} model uncertainty are challenging to execute with high dimensional raw sensory input (e.g. images).
More importantly, the agent should learn to deal with `stochasticity' in its interaction with the environment caused by either noisy actuation of the agent's motors, or the observations could be inherently stochastic.
A deterministic prediction model will always end up with a non-zero prediction error allowing the agent to get stuck in the local minima of exploration.

Similar behavior would occur if the task at hand is too difficult to learn. Consider a robotic arm manipulating a keybunch. Predicting the change in pose and position of each key in the keybunch is extremely difficult. Although the behavior is not inherently stochastic, our agent could easily get stuck in playing with the same keybunch and not try other actions or even other objects. Existing formulations of curiosity reward or novelty-seeking count-based methods would also suffer in such scenarios. Learning probabilistic predictive models to measure uncertainty~\cite{houthooft2016vime}, or measuring learnability by capturing the change in prediction error~\cite{schmidhuber1991curious,oudeyer2009intrinsic} have been proposed as solutions, but have been demonstrated in low-dimensional state space inputs and are difficult to scale to high dimensional image inputs.

\subsection{Disagreement as Intrinsic Reward}
\label{sec:formulation}
Instead of learning a single dynamics model, we propose an alternate exploration formulation based on ensemble of models as inspired by the classical active learning literature~\cite{Seung1992}.
The goal of active learning is to find the optimal training examples to label such that the accuracy is maximized at minimum labeling cost.
While active learning minimizes optimal cost with an analytic policy, the goal of an exploration-driven agent is to learn a policy that allows it to best navigate the environment space.
Although the two might look different at the surface, we argue that active learning objectives could inspire powerful intrinsic reward formulations. In this work, we leverage the idea of model-variance maximization to propose exploration formulation.
Leveraging model variance to investigate a system is also a well-studied mechanism in optimal experimental design literature~\cite{boyd2004convex} in statistics.

As our agent interacts with the environment, it collects trajectory of the form $\{x_t, a_t, x_{t+1}\}$.
After each rollout, the collected transitions are used to train an ensemble of forward prediction models $\{ f_{\theta_1}, f_{\theta_2} \dots, f_{\theta_k}\}$ of the environment.
Each of the model is trained to map a given tuple of current observation $x_t$ and the action $a_t$ to the resulting state $x_{t+1}$.
These models are trained using straightforward maximum likelihood estimation that minimizes the prediction error, i.e, $\|f(x_t, a_t; \theta) - x_{t+1}\|_2$.
To maintain the diversity across the individual models, we initialize each model's parameters differently and train each of them on a subset of data randomly sampled with replacement (bootstrap).

Each model in our ensemble is trained to predict the ground truth next state. Hence, the parts of the state space which have been well explored by the agent will have gathered enough data to train all models, resulting in an agreement between the models. Since the models are learned (and not tabular), this property should generalize to unseen but similar parts of the state-space. However, the areas which are novel and unexplored would still have high prediction error for all models as none of them are yet trained on such examples, resulting in disagreement on the next state prediction. Therefore, we use this disagreement as an intrinsic reward to guide the policy. Concretely, the intrinsic reward $r^i_t$ is defined as the \textit{variance} across the output of different models in the ensemble:
\begin{align}
    r^i_t \triangleq \mathbb{E}_{\theta} \Big[ \|f(x_t, a_t; \theta) - \mathbb{E}_{\theta}[f(x_t, a_t; \theta)]\|_2^2 \Big] \label{eq:rewardint}
\end{align}
Note that the expression on the right does not depend on the next state $x_{t+1}$ --- a property which will exploit in Section~\ref{sec:differentiable} to propose efficient policy optimization.

Given the agent's rollout sequence and the intrinsic reward $r^i_t$ at each timestep $t$, the policy is trained to maximize the sum of expected reward, i.e., $\max_{\theta_P} \mathbb{E}_{\pi(x_t; \theta_P)}\big[\sum_t \gamma^t r^i_t\big]$ discounted by a factor $\gamma$. Note that the agent is self-supervised and does not need any extrinsic reward to explore.
The agent policy and the forward model ensemble are jointly trained in an online manner on the data collected by the agent during exploration.
This objective can be maximized by any policy optimization technique, e.g., we use proximal policy optimization (PPO)~\cite{ppo} unless specified otherwise.

\subsection{Exploration in Stochastic Environments}
Consider a scenario where the next state $x_{t+1}$ is stochastic with respect to the current state $x_t$ and action $a_t$. The source of stochasticity could be noisy actuation, difficulty or inherent randomness. Given enough samples, a dynamic prediction model should learn to predict the mean of the stochastic samples. Hence, the variance of the outputs in ensemble will drop preventing the agent from getting stuck in stochastic local-minima of exploration. Note this is unlike prediction error based objectives~\cite{pathakICMl17curiosity,schmidhuber1991possibility} which will settle down to a mean value after large enough samples. Since, the mean is different from the individual ground-truth stochastic states, the prediction error remains high making the agent forever curious about the stochastic behavior. We empirically verify this intuition by comparing prediction-error to disagreement across several environments in Section~\ref{sec:stochasticExp}.

\subsection{Differentiable Exploration for Policy Optimization}
\label{sec:differentiable}
One commonality between different exploration methods~\cite{bellemare2016unifying,pathakICMl17curiosity,houthooft2016vime}, is that the prediction model is usually learned in a supervised manner and the agent's policy is trained using reinforcement learning either in on-policy or off-policy manner.
Despite several formulations over the years, the policy optimization procedure to maximize these intrinsic rewards has more or less remained the same -- i.e. -- treating the intrinsic reward as a ``black-box'' even though it is generated by the agent itself.

Let's consider an example to understand the reason behind the status quo.
Consider a robotic-arm agent trying to push multiple objects kept on the table in front of it by looking at the image from an overhead camera.
Suppose the arm pushes an object such that it collides with another one on the table. The resulting image observation will be the outcome of complex real-world interaction, the actual dynamics of which is not known to the agent.
Note that this resulting image observation is a function of the agent's action (i.e., push in this case).
Most commonly, the intrinsic reward $r^i(x_t, a_t, x_{t+1})$ is function of the next state (which is a function of the agent's action), e.g., information gain~\cite{houthooft2016vime}, prediction error~\cite{pathakICMl17curiosity} etc.
This dependency on the unknown environment dynamics absolves the policy optimization of analytical reward gradients with respect to the action. Hence, the standard way is to optimize the policy to maximize the sequence of intrinsic rewards using reinforcement learning, and not make any use of the structure present in the design of $r^i_t$.

We formulate our proposed intrinsic reward as a differentiable function so as to perform policy optimization using likelihood maximization -- much like supervised learning instead of reinforcement.
If possible, this would allow the agent to make use of the structure in $r^i_t$ explicitly, i.e., the intrinsic reward from the model could very efficiently inform the agent to change its action space in the direction where forward prediction loss is high, instead of providing a \textit{scalar} feedback as in case of reinforcement learning.
Explicit reward (cost) functions are one of the key reasons for success stories in optimal-control based robotics~\cite{deisenroth2011pilco,gal2016improving}, but they don't scale to high-dimensional state space such as images and rely on having access to a good model of the environment.

We first discuss the one step case and then provide the general setup.
Note that our intrinsic reward formulation, shown in Equation~\eqref{eq:rewardint}, does not depend on the environment interaction at all, i.e., no dependency on $x_{t+1}$. It is purely a mental simulation of the ensemble of models based on the current state and the agent's prediction action.
Hence, instead of maximizing the intrinsic reward in expectation via PPO (RL), we can optimize for policy parameters $\theta_P$ using direct gradients by treating $r^i_t$ as a differentiable loss function. The objective for a one-step reward horizon is:
\begin{align}
    \min_{\theta_1,\dots,\theta_k}\;\; & (1/k)\sum_{i=1}^k \|f_{\theta_i}(x_t, a_t) - x_{t+1}\|_2\label{eq:jointonestep} \\
    \max_{\theta_P}\;\; & (1/k)\sum_{i=1}^k \Big[ \|f_{\theta_i}(x_t, a_t) - (1/k)\sum_{j=1}^k f_{\theta_j}(x_t, a_t)\|_2^2 \Big] \nonumber\\
    \text{s.t.}\;\; & a_t = \pi({x}_t; \theta_P) \nonumber
\end{align}
This is optimized in an alternating fashion where the forward predictor is optimized keeping the policy parameters frozen and vice-versa.
Note that both policy and forward models are trained via maximum likelihood in a supervised manner, and hence, efficient in practice.

\paragraph{Generalization to multi-step reward horizon} To optimize policy for maximizing a discounted sum of sequence of future intrinsic rewards $r_t^i$ in a differentiable manner, the forward model would have to make predictions spanning over multiple time-steps. The policy objective in Equation~\eqref{eq:jointonestep} can be generalized to the multi-step horizon setup by recursively applying the forward predictor, i.e., $\max_{\theta_P}\sum_t r^i_t(\hat{x}_t, a_t)$ where $\hat{x}_{t} = f(\hat{x}_{t-1}, a_{t-1}; \theta)$, $a_t = \pi({x}_t; \theta_P)$, $\hat{x}_{0} = x_{0}$, and $r^i_t(.)$ is defined in Equation~\eqref{eq:rewardint}. Alternatively, one could use LSTM to make forward model itself multi-step. However, training a long term multi-step prediction model is challenging and an active area of research.
In this paper, we show differentiable exploration results for short horizon only and leave multi-step scenarios for future work.

\section{Implementation Details and Baselines}
\label{sec:details}
\paragraph{Learning forward predictions in the feature space} It has been shown that learning forward-dynamics predictor $f_{\theta}$ in a feature space leads to better generalization in contrast to raw pixel-space predictions~\cite{burda2018large,pathakICMl17curiosity}. Our formulation is trivially extensible to any representation space $\phi$ because all the operations can be performed with $\phi(x_t)$ instead of $x_t$. Hence, in all of our experiments, we train our forward prediction models in feature space. In particular, we use random feature space in all video games and navigation, classification features in MNIST and ImageNet-pretrained ResNet-18 features in real world robot experiments. We use 5 models in the ensemble.

\paragraph{Back-propagation through forward model} To directly optimize the policy with respect to the loss function of the forward predictor, as discussed in Section~\ref{sec:differentiable}, we need to backpropagate all the way through action sampling process from the policy. In case of continuous action space, one could achieve this via making policy deterministic, i.e. $a_t = \pi_{\theta_P}$ with epsilon-greedy sampling~\cite{lillicrap2015continuous}. For discrete action space, we found that straight-through estimator~\cite{bengio2013estimating} works well in practice.

\begin{figure*}[t]
\vspace{-0.1in}
\centering
    \includegraphics[width=1.0\linewidth]{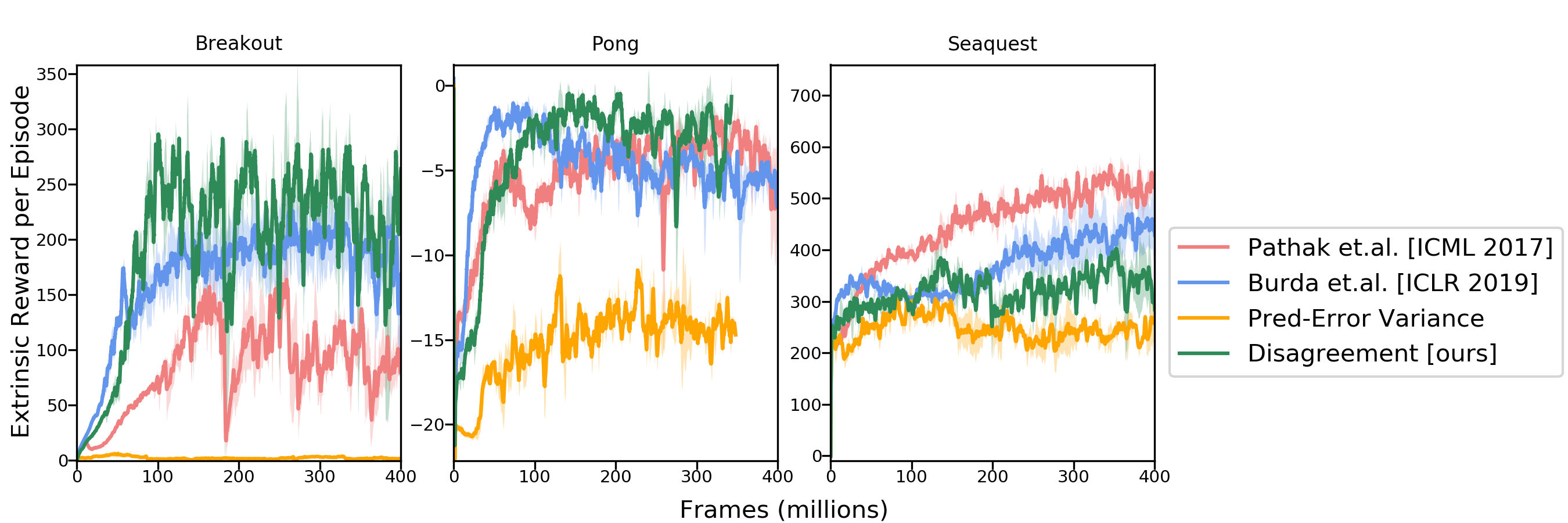}
    \vspace{-0.25in}
    \caption{\textbf{Sanity Check in Non-Stochastic Environments}: We compare different intrinsic reward formulations across near-deterministic, non-stochastic standard benchmark of the Atari games. Our disagreement-based approach compares favorably to state-of-the-art approaches without losing accuracy in non-stochastic scenarios.}
    \label{fig:sanity_check}
\vspace{-0.15in}
\end{figure*}

\paragraph{Baseline Comparisons}
`Disagreement' refers to our exploration formulation optimized using PPO~\cite{ppo} as discussed in Section~\ref{sec:formulation}, unless mentioned otherwise. `Disagreement [Differentiable]' refers to the direct policy optimization for our formulation as described in Section~\ref{sec:differentiable}. `Pathak et.al. [ICML 2017]' refers to the curiosity-driven exploration formulation based on the prediction error of the learned forward dynamics model in inverse model action space~\cite{pathakICMl17curiosity}. `Burda et.al. [ICLR 2019]' refers to the random feature-based prediction-error~\cite{burda2018large}. `Pred-Error Variance' is an alternative ablation where we train the agent to maximize the variance of the prediction error as opposed to the variance of model output itself.
Finally, we also compare our performance to Bayesian Neural Networks for measuring variance. In particular, we compared to Dropout NN~\cite{gal2015} represented as `Bayesian Disagreement'.

\section{Experiments}
We evaluate our approach on several environments including Atari games, 3D navigation in Unity, MNIST, object manipulation in Mujoco and real world robotic manipulation task using Sawyer arm.
Our experiments comprise of three parts: a) verifying the performance on standard non-stochastic environments; b) comparison on environments with stochasticity in either transition dynamics or observation space; and c) validating the efficiency of differentiable policy optimization facilitated by our objective.

\subsection{Sanity Check in Non-Stochastic Environments}
\label{sec:atari}
We first verify whether our disagreement formulation is able to maintain the performance on the standard environment as compared to state of the art exploration techniques. Although the primary advantage of our approach is in handling stochasticity and improving efficiency via differentiable policy optimization, it should not come at the cost of performance in nearly-deterministic scenarios. We run this sanity check on standard Atari benchmark suite, as shown in Figure~\ref{fig:sanity_check}.
These games are not completely deterministic and have some randomness as to where the agent is spawned upon game resets~\cite{dqn}.
The agent is trained with only an intrinsic reward, without any external reward from the game environment. The external reward is only used as a proxy to evaluate the quality of exploration and not shown to the agent.

We train our ensemble of models for computing disagreement in the embedding space of a random network as discussed in Section~\ref{sec:details}.
The performance is compared to curiosity formulation~\cite{pathakICMl17curiosity}, curiosity with random features~\cite{burda2018large}, Bayesian network based uncertainty and variance of prediction error.
As seen in the results, our method is as good as or slightly better than state-of-the-art exploration methods in most of the scenarios. Overall, these experiments suggest that our exploration formulation which is only driven by disagreement between models output compares favorably to state of the art methods.
Note that the variance of prediction error performs significantly worse.
This is so because the low variance in prediction error of different models doesn't necessarily mean they will agree on the next state prediction. Hence, `Pred-Error Variance' may sometimes incorrectly stop exploring even if output prediction across models is drastically different.

\subsection{Exploration in Stochastic Environments}
\label{sec:stochasticExp}

\paragraph{A) Noisy MNIST.}
We first build a toy task on MNIST to intuitively demonstrate the contrast between disagreement-based intrinsic reward and prediction error-based reward~\cite{pathakICMl17curiosity} in stochastic setups.
This is a one-step environment where the agent starts by randomly observing an MNIST image from either class 0 or class 1.
The dynamics of the environment are defined as follows: 1) images with label 0 always transition to another image from class 0. 2) Images with label 1 transition to a randomly chosen image from class label 2 to 9. This ensures that a transition from images with label 0 has low stochasticity (i.e., transition to the same label). On the other hand, transitions from images with label 1 have high stochasticity. The ideal intrinsic reward function should give similar incentive (reward) to both the scenarios after the agent has observed a significant number of transitions.
\begin{figure}
\vspace{-0.1in}
\centering
    \includegraphics[width=0.75\linewidth]{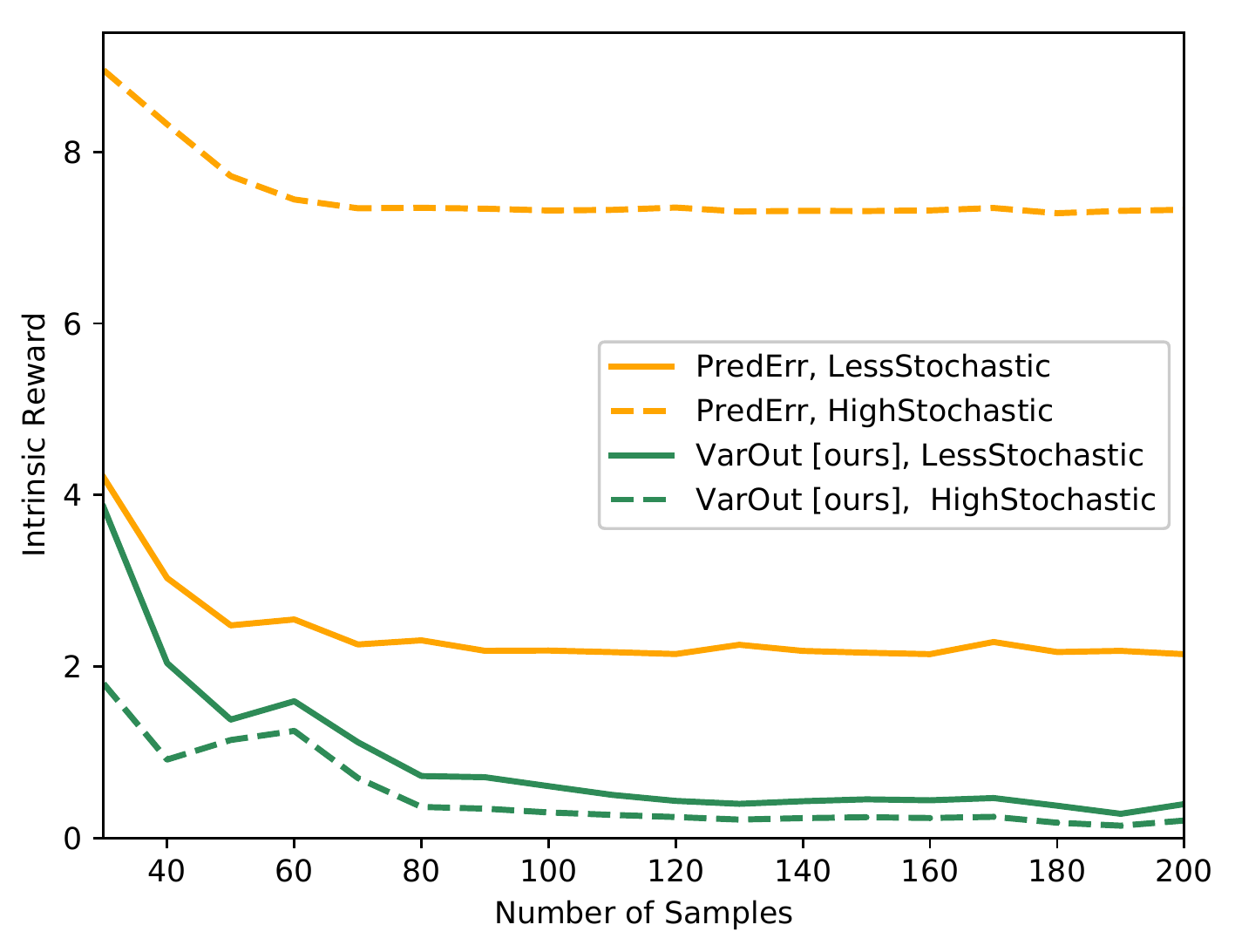}
      \vspace{-0.15in}
    \caption{Performance of disagreement across ensemble vs prediction error based reward function on Noisy MNIST environment. This environment has 2 sets of state with different level of stochasticity associated with them. The disagreement-based intrinsic reward converges to the ideal case of assigning the same reward value for both states. However, the prediction-error based reward function assigns a high reward to states with high stochasticity. }
    \label{fig:mnist}
    \vspace{-0.15in}
\end{figure}

Figure~\ref{fig:mnist} shows the performance of these methods on the test set of MNIST as a function of the number of states visited by the agent.
Even at convergence, the prediction error based model assigns more reward to the observations with higher stochasticity, i.e., images with label 1. This behavior is detrimental since the transition from states of images with label 1 cannot ever be perfectly modeled and hence the agent will get stuck forever.
In contrast, our ensemble-based disagreement method converges to almost zero intrinsic reward in both the scenarios after the agent has seen enough samples, as desired.

\begin{figure}
\vspace{-0.1in}
\centering
    \includegraphics[width=\linewidth]{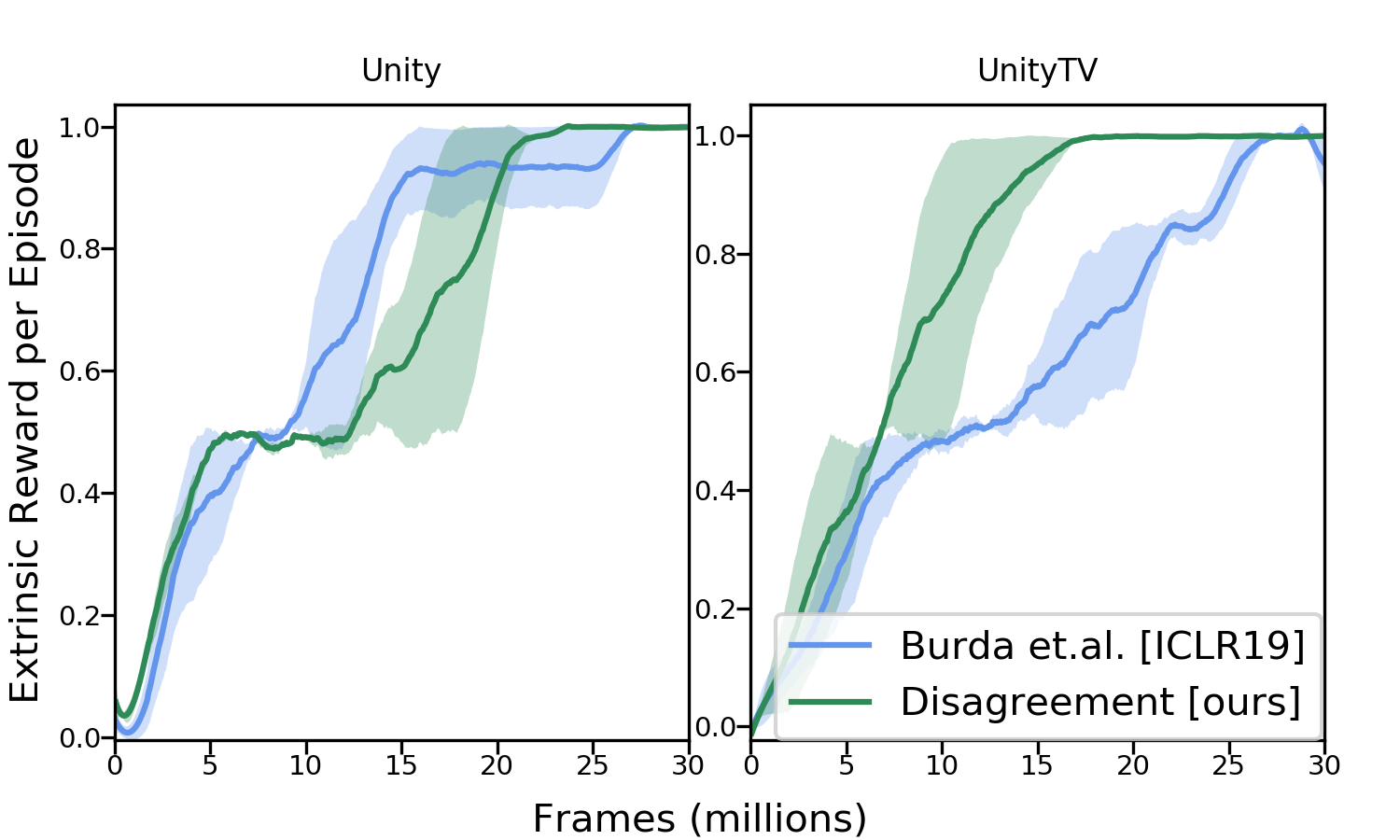}
    \vspace{-0.25in}
    \caption{\textbf{3D Navigation in Unity}: Comparison of prediction-error based curiosity reward with our proposed disagreement-based exploration on 3D navigation task in Unity with and without the presence of TV+remote. While both the approaches perform similar in normal case (left), disagreement-based approach performs better in the presence of stochasticity (right).}
    \label{fig:unityTv}
    \vspace{-0.15in}
\end{figure}

\paragraph{B) 3D Navigation in Unity.}
The goal in this setup is to train the agent to reach a target location in the maze. The agent receives a sparse reward of +1 on reaching the goal. For all the methods, we train the policy of the agent to maximize the summation of intrinsic and sparse extrinsic reward. This particular environment is a replica of VizDoom-MyWayHome environment in unity ML-agent and was proposed in~\citet{burda2018large}. Interestingly, this environment has 2 variants, one of which has a TV on the wall. The agent can change the channel of the TV but the content is stochastic (random images appear after pressing button). The agent can start randomly anywhere in the maze in each episode, but the goal location is fixed. We compare our proposed method with state-of-the-art prediction error-based exploration~\cite{burda2018large}. The results are shown in Figure~\ref{fig:unityTv}. Our approach performs similar to the baseline in the non-TV setup and outperforms the baseline in the presence of the TV. This result demonstrates that an ensemble-based disagreement could be a viable alternative in realistic stochastic setups.

\begin{figure}
\vspace{-0.12in}
\centering
    \includegraphics[width=\linewidth]{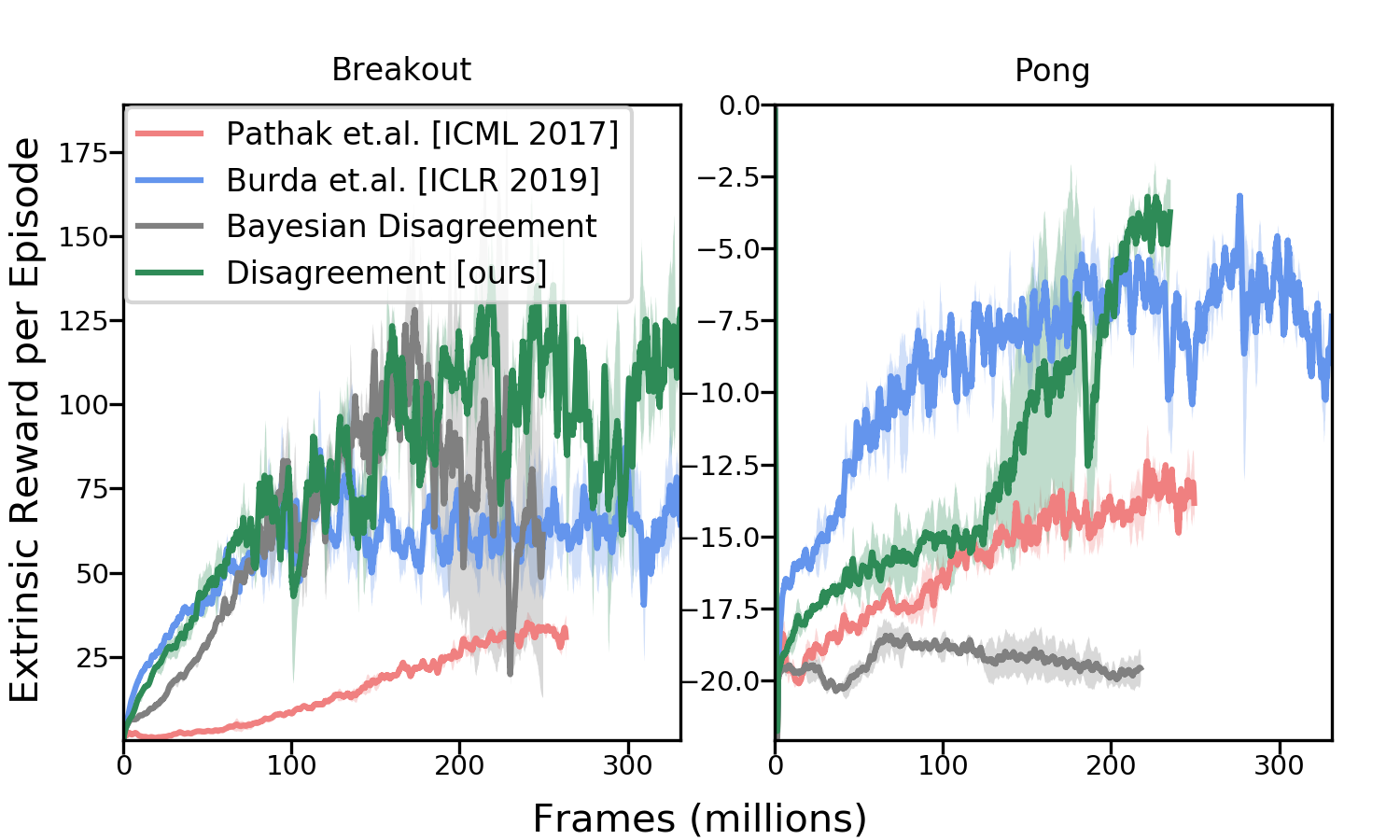}
        \vspace{-0.25in}
    \caption{\textbf{Stochastic Atari Games}: Comparison of different exploration techniques in the the Atari (`sticky') environment. The disagreement-based exploration is robust across both the scenarios.}
      \vspace{-0.2in}
    \label{fig:sticky_atari}
\end{figure}

\paragraph{C) Atari with Sticky Actions.}
As discussed in Section~\ref{sec:atari}, the usual Atari setup is nearly deterministic. Therefore, a recent study~\cite{stickyAtari} proposed to introduce stochasticity in Atari games by making actions `sticky', i.e., at each step, either the agent's intended action is executed or the previously executed action is repeated with equal probability.
As shown in Figure~\ref{fig:sticky_atari}, our disagreement-based exploration approach outperforms previous state-of-the-art approaches. In Pong, our approach starts slightly slower than Burda et.al.~\cite{burda2018large}, but eventually achieves a higher score.
Further note that the Bayesian network-based disagreement does not perform as well as ensemble-based disagreement. This suggests that perhaps dropout~\cite{gal2015} isn't able to capture good uncertainty estimate in practice.
These experiments along with the navigation experiment, demonstrate the potential of ensembles in the face of stochasticity.

\subsection{Differentiable Exploration in Structured Envs}
We now evaluate the differentiable exploration objective proposed in Section~\ref{sec:differentiable}. As discussed earlier, the policy is optimized via direct analytic gradients from the exploration module.  Therefore, the horizon of exploration depends directly on the horizon of the module. Since training long-horizon models from high dimensional inputs (images) is still an unsolved problem, we evaluate our proposed formulation on relatively short horizon scenarios. However, to compensate for the length of the horizon, we test on large action space setups for real-world robot manipulation task.

\begin{figure}
\vspace{-0.1in}
\centering
    \includegraphics[width=0.8\linewidth]{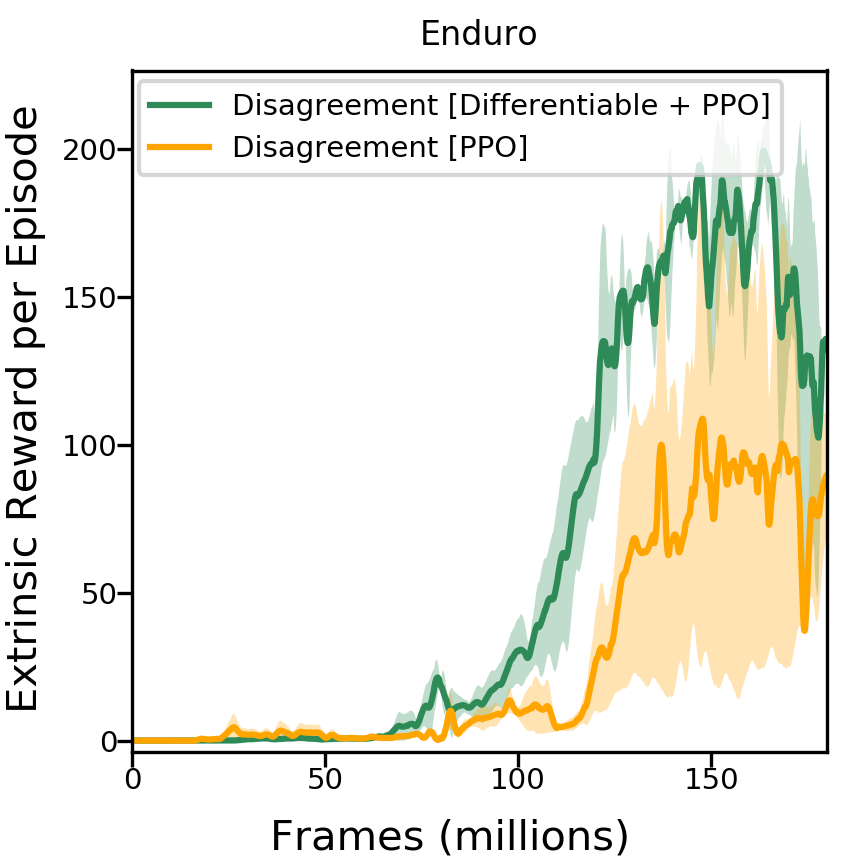}
    \vspace{-0.1in}
    \caption{Performance comparison of disagreement-based exploration with or without the differentiable policy optimization in Enduro Atari Game. Differentiability helps the agent learn faster.}
    \vspace{-0.15in}
    \label{fig:enduro}
\end{figure}

\paragraph{A) Enduro Video Game.}
In this game, the goal of the agent is to steer the car on racing track to avoid enemies. The agent is trained to explore via purely intrinsic rewards, and the extrinsic reward is only used for evaluation. In order to steer the car, the agent doesn't need to model long-range dependencies. Hence, in this environment, we combine our differentiable policy optimization with reinforcement learning (PPO) to maximize our disagreement based intrinsic reward. The RL captures discounted long term dependency while our differentiable formulation should efficiently take care of short-horizon dependencies. We compare this formulation to purely PPO based optimization of our intrinsic reward. As shown in Figure~\ref{fig:enduro}, our differentiable exploration expedites the learning of the agent suggesting the efficacy of direct gradient optimization. We now evaluate the performance of only differentiable exploration (without reinforcement) in short-horizon and large-structured action space setups.

\subsection*{B) Object Manipulation by Exploration.}
We consider the task of object manipulation in complex scenarios. Our setup consists of a 7-DOF robotic arm that could be tasked to interact with the objects kept on the table in front of it.
The objects are kept randomly in the workspace of the robot on the table.  Robot's action space is end-effector position control: a) location $(x, y)$ of point on the surface of table, b) angle of approach $\theta$, and c) gripper status, a binary value indicating whether to grasp (open the gripper fingers) or push (keep fingers close). All of our experiments use raw visual RGBD images as input and predict actions as output. Note that, to accurately grasp or push objects, the agent needs to figure out an accurate combination of location, orientation and gripper status.

The action space is discretized into $224\times224$ locations, 16 orientations for grasping (fingers close) and 16 orientations for pushing leading to final dimension of $224\times224\times32$.
The policy takes as input a $224\times224$ RGBD image and produces push and grasp action probabilities for each pixel. Following~\cite{DBLP:journals/corr/abs-1803-09956}, instead of adding the 16 rotations in the output, we pass 16 equally spaced rotated images to the network and then sample actions based on the output of all the inputs. This exploits the convolutional structure of the network. The task has a short horizon but very large state and action spaces. We make no assumption about either the environment or the training signal. Our robotic agents explore the work-space purely out of their own intrinsic reward in a pursuit to develop useful skills. We have instantiated this setup in a Mujoco simulation as well as in the real world robotics scenarios.

\begin{figure*}[t!]
\centering
\vspace{-0.1in}
\begin{subfigure}[b]{0.32\linewidth}
    \includegraphics[width=0.9\linewidth]{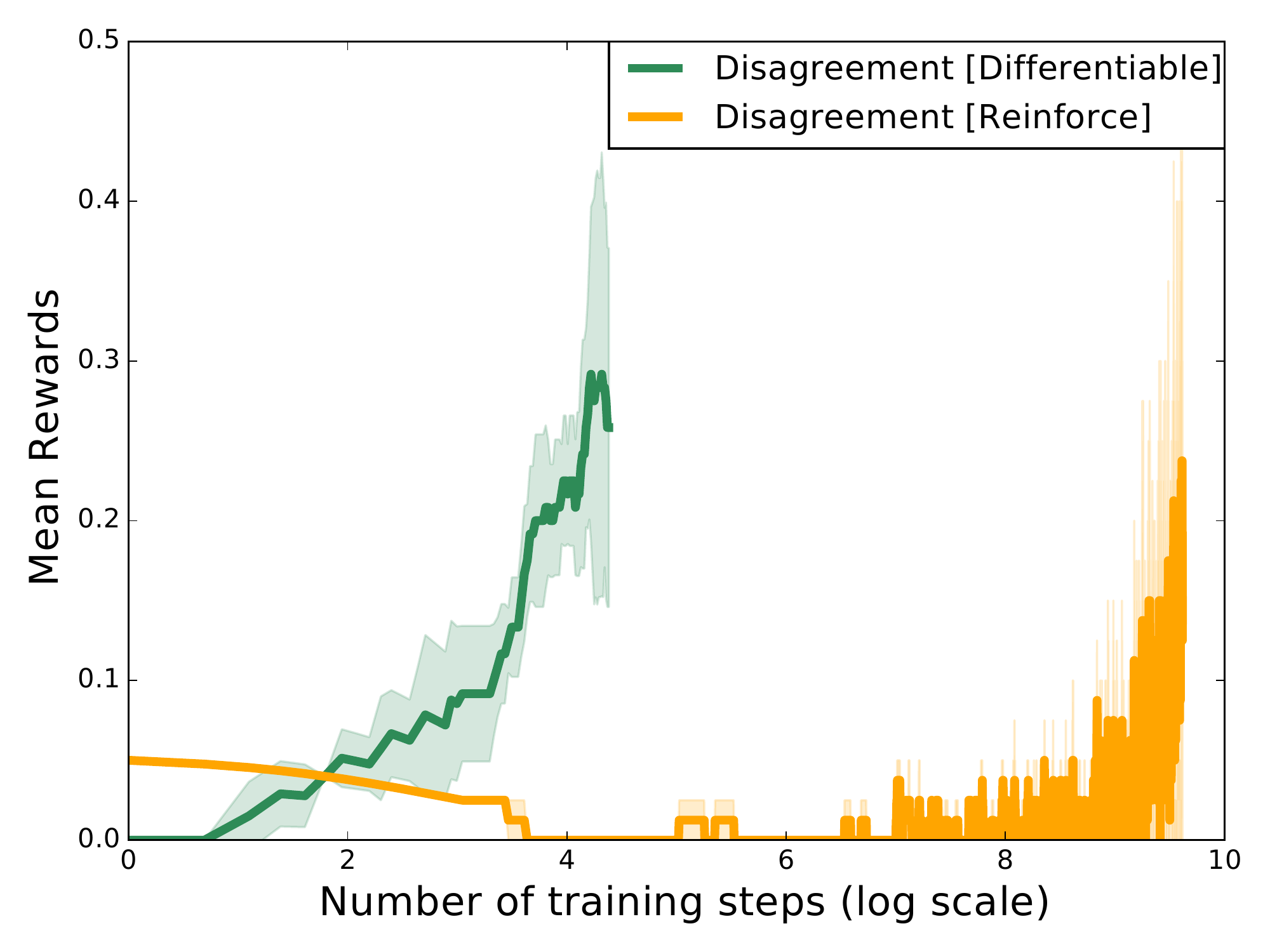}
    \caption{Mujoco}
    \label{fig:mujoco}
\end{subfigure}
\begin{subfigure}[b]{0.3\linewidth}
    \includegraphics[width=0.9\linewidth]{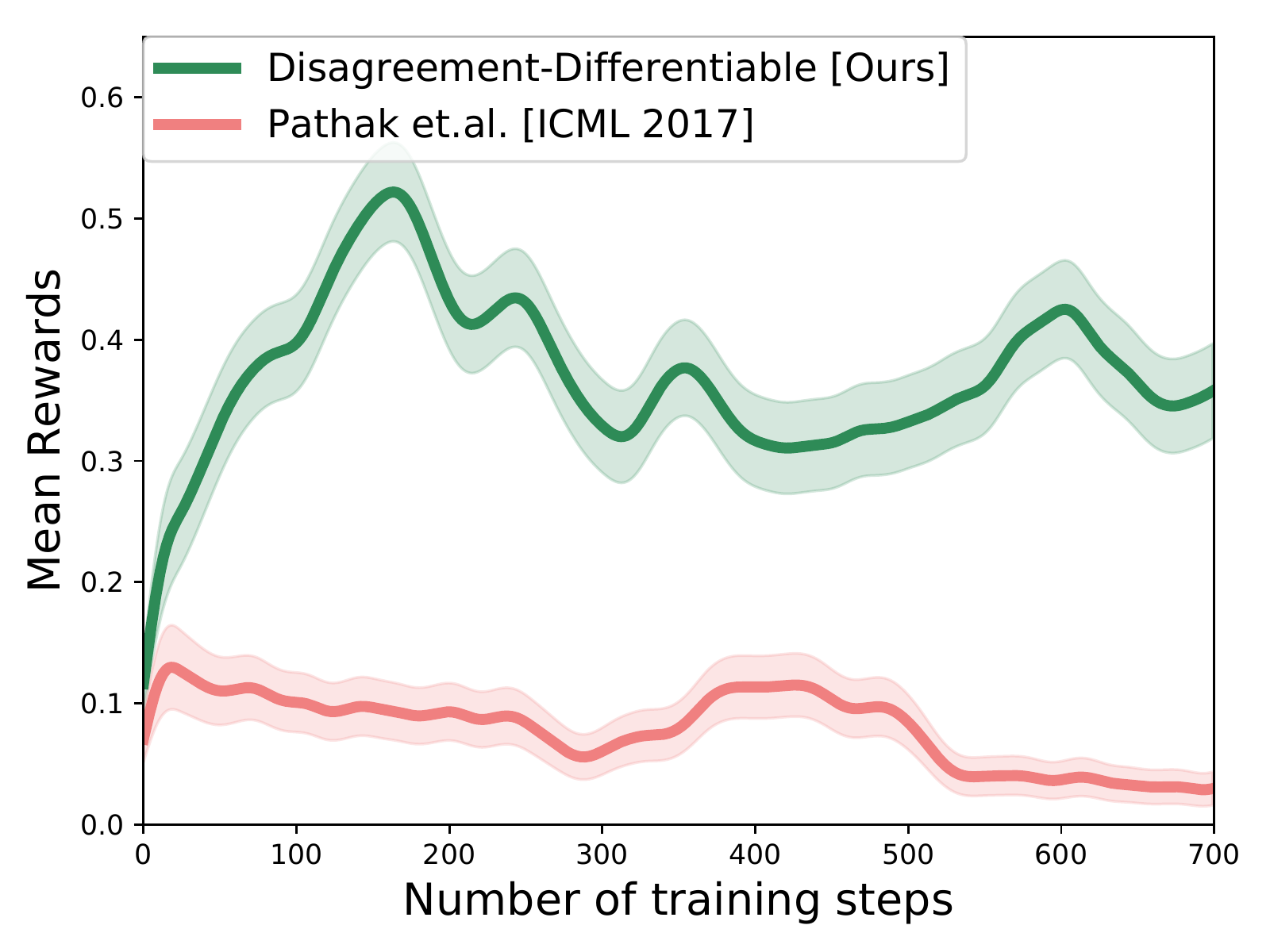}
    \caption{Real Robot}
    \label{fig:real_multiple11}
\end{subfigure}
\begin{subfigure}[b]{0.3\linewidth}
    \includegraphics[width=0.9\linewidth]{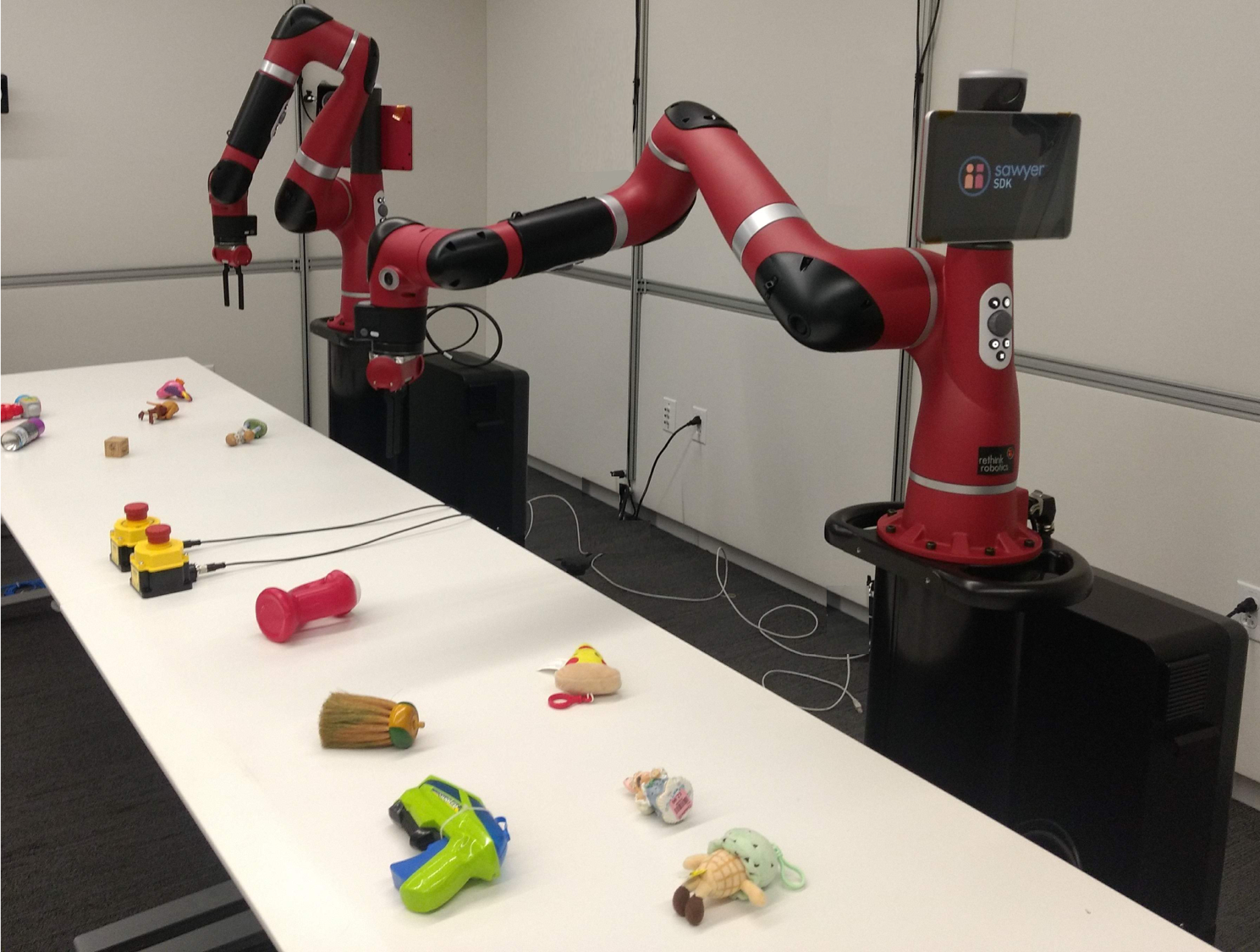}
    \caption{Real Robot Setup}
    \label{fig:realRobot}
\end{subfigure}
\vspace{-0.15in}
\caption{Measuring object interaction rate with respect to the number of samples in (a) Mujoco, and (b) real-world robot. Note that the Mujoco plot is in log-scale. We measure the exploration quality by evaluating the object interaction frequency of the agent. In both the environments, our differentiable policy optimization explores more efficiently. (c) A snapshot of the real-robotic setup.
}
\label{fig:manipulation}
\vspace{-0.1in}
\end{figure*}

\paragraph{B1) Object Manipulation in MuJoCo.}
We first carry out a study in simulation to compare the performance of differentiable variant of our disagreement objective against the reinforcement learning based optimization. We used MuJoCo to simulate the robot performing grasping and pushing on tabletop environment as described above.

To evaluate the quality of exploration, we measure the frequency at which our agent interacts (i.e., touches) with the object. This measure is just used to evaluate the exploration quantitatively and is not used as a training signal. It represents how quickly our agent's policy learns to explore an interesting part of space. Figures~\ref{fig:mujoco} shows the performance when the environment consists of just a single object which makes it really difficult to touch the object randomly. Our approach is able to exploit the structure in the loss, resulting in order of magnitude faster learning than REINFORCE.

\paragraph{B2) Real-World Robotic Manipulation.}
We now deploy our sample-efficient exploration formulation on real-world robotics setup. The real-world poses additional challenges, unlike simulated environments in terms of behavior and the dynamics of varied object types.
Our robotic setup consisted of a Sawyer-arm with a table placed in front of it. We mounted KinectV2 at a fixed location from the robot to receive RGBD observations of the environment.

In every run, the robot starts with 3 objects placed in front of it. Unlike other self-supervised robot learning setups, we keep fewer objects to make exploration problem harder so that it is not trivial to interact with the objects by acting randomly. If either the robot completes 100 interactions or there are no objects in front of it, objects are replaced manually. Out of a total of 30 objects, we created a set of 20 objects for training and 10 objects for testing. We use the same metric as used in the simulation above (i.e., number of object interactions) to measure the effectiveness of our exploration policy during training. We monitor the change in the RGBD image to see if the robot has interacted with objects. Figure~\ref{fig:real_multiple11} shows the effectiveness of differentiable policy optimization for disagreement over prediction-error based curiosity objective. Differentiable-disagreement allows the robotic agent to learn to interact with objects in less than 1000 examples.

We further test the skills learned by our robot during its exploration by measuring object-interaction frequency on a set of 10 held-out test objects.
For both the methods, we use the checkpoint saved after 700 robot interaction with the environment.
For each model, we evaluate a total of 80 robot interaction steps with three test objects kept in front. The environment is reset after every 10 robot steps during evaluation.
Our final disagreement exploration policy interacts approximately \textbf{67\%} of times with unseen objects, whereas a random policy performs at \textbf{17\%}. On the other hand, it seems that REINFORCE-based curiosity policy just collapses and only \textbf{1\%} of actions involve interaction with objects. Videos are available at~\url{https://pathak22.github.io/exploration-by-disagreement/}.

\section{Related Work}
Exploration is a well-studied problem in the field of reinforcement learning. Early approaches focused on studying exploration from theoretical perspective~\cite{strehl08} and proposed Bayesian formulations~\cite{kolter09,Deisenroth11} but they are usually hard to scale to higher dimensions (e.g., images). In this paper, we focus on the specific problem of exploration using intrinsic rewards. A large family of approaches use ``curiosity'' as an intrinsic reward for training the agents. A good summary of early work in curiosity-driven rewards can be found in ~\cite{oudeyer2007intrinsic,oudeyer2009intrinsic}. Most approaches use some form of prediction-error between the learned model and environment behavior~\cite{pathakICMl17curiosity}. This prediction error can also be formulated as surprise~\cite{schmidhuber1991curious, josh_surprise, sun2011planning}. Other techniques incentivize exploration of states and actions where prediction of a forward model is highly-uncertain~\cite{still2012information, houthooft2016vime}. Finally, approaches such as~\citet{lopes2012exploration} try to explore state space which help improve the prediction model. Please refer to the introduction Section~\ref{sec:intro} for details on formulations using curiosity, visitation count or diversity. However, most of these efforts study the problem in the context of external rewards.

Apart from intrinsic rewards, other approaches include using an adversarial game~\cite{sukhbaatar2017intrinsic} where one agent gives the goal states and hence guiding exploration. \citet{gregor2017variational} introduce a formulation of empowerment where agent prefers to go to states where it expects it will achieve the most control after learning.
Researchers have also tried using perturbation of learned policy for exploration~\cite{fortunato2017noisy,fu2017ex2,plappert2017parameter} and using value function estimates~\cite{osband2016deep}. Again these approaches have mostly been considered in the context of external rewards and are not efficient enough to be scalable to real robotics setup.

Our work is inspired by large-body of work in active learning (AL). In the AL setting, given a collection of unlabeled examples, a learner selects which samples will be labeled by an oracle~\cite{Settles2010}. Common selection criteria include entropy~\cite{Dagan1995}, uncertainty sampling~\cite{Lewis1994} and expected informativeness~\cite{Houlsby2011}.
Our work is inspired by by~\cite{Seung1992}, and we apply the disagreement idea in a completely different setting of exploration and show its applicability to environments with stochastic dynamics and improving sample-efficiency.
Concurrent to this work,~\citet{shyam2019max} also show the effectiveness of model-based exploration in estimating novelty, and~\citet{henaff2019model} use variance regularization for policy learning via imitation.

\section*{Acknowledgements}
We would like to thank Ben Recht, Leon Bottou, Harri Edwards, Yuri Burda, Ke Li, Saurabh Gupta, Shubham Tulsiani, and Yann Lecun for fruitful discussions and comments. Part of the work was performed when DP was interning at Facebook AI Research. DP is supported by the Facebook graduate fellowship.

\bibliography{main}
\bibliographystyle{icml2019}

\end{document}